%% file: green.tex
\pdfoutput=1
\documentclass[11pt]{article}

\usepackage{acl}
\usepackage{times}
\usepackage{latexsym}
\usepackage{booktabs}
\usepackage[T1]{fontenc}
\usepackage[utf8]{inputenc}
\usepackage{microtype}
\usepackage{graphicx}
\usepackage{todonotes}
\usepackage{amsmath}
\usepackage{xurl}
\usepackage{soul}
\usepackage{adjustbox}
\usepackage{paralist}
\usepackage{fancyvrb}
\usepackage{listings}
\title{Towards Climate Awareness in NLP Research} 

\author{Daniel Hershcovich \\
  Department of Computer Science \\
  University of Copenhagen \\
  \texttt{dh@di.ku.dk} \\\And
  Nicolas Webersinke \quad Mathias Kraus \\
  FAU Erlangen-Nuremberg \\
  \texttt{\{nicolas.webersinke,}\\\texttt{mathias.kraus\}@fau.de} \\\AND
  Julia Anna Bingler \\
  ETH Zurich \\
  \texttt{binglerj@ethz.ch} \\\And
  Markus Leippold \\
  University of Zurich \\
  \texttt{markus.leippold@bf.uzh.ch}}

\begin{document}
\maketitle
\begin{abstract}
The climate impact of AI, and NLP research in particular, has become a serious issue given the enormous amount of energy that is increasingly being used for training and running computational models. Consequently, increasing focus is placed on efficient NLP. However, this important initiative lacks simple guidelines that would allow for systematic climate reporting of NLP research. We argue that this deficiency is one of the reasons why very few publications in NLP report key figures that would allow a more thorough examination of environmental impact, and present a quantitative survey to demonstrate this. As a remedy, we propose a climate performance model card with the primary purpose of being practically usable with only limited information about experiments and the underlying computer hardware. We describe why this step is essential to increase awareness about the environmental impact of NLP research and, thereby, paving the way for more thorough discussions.\footnote{We provide a Jupyter notebook with the code used to conduct our survey, as well as model card templates in {\LaTeX} and Markdown, at \url{https://github.com/danielhers/climate-awareness-nlp}.}
\end{abstract}

\section{Introduction}\label{sec:intro}

As Artificial Intelligence (AI), and specifically Natural Language Processing (NLP), scale up to require more computational resources and thereby more energy, there is an increasing focus on efficiency and sustainability \cite{strubell-etal-2019-energy,Schwartz:2020}.
For example, training a single BERT base model \cite{devlin-etal-2019-bert} requires as much
energy as a trans-American flight \cite{strubell-etal-2019-energy}.
While newer models are arguably more efficient \cite{fedus2021switch,borgeaud2022improving,so2022primer}, they are also an order of magnitude larger, raising environmental concerns \cite{bender2021dangers}. The problem will only worsen with time, as compute requirements double every 10 months \cite{sevilla2022compute}.

\begin{figure}
    \centering
    \includegraphics[width=\columnwidth]{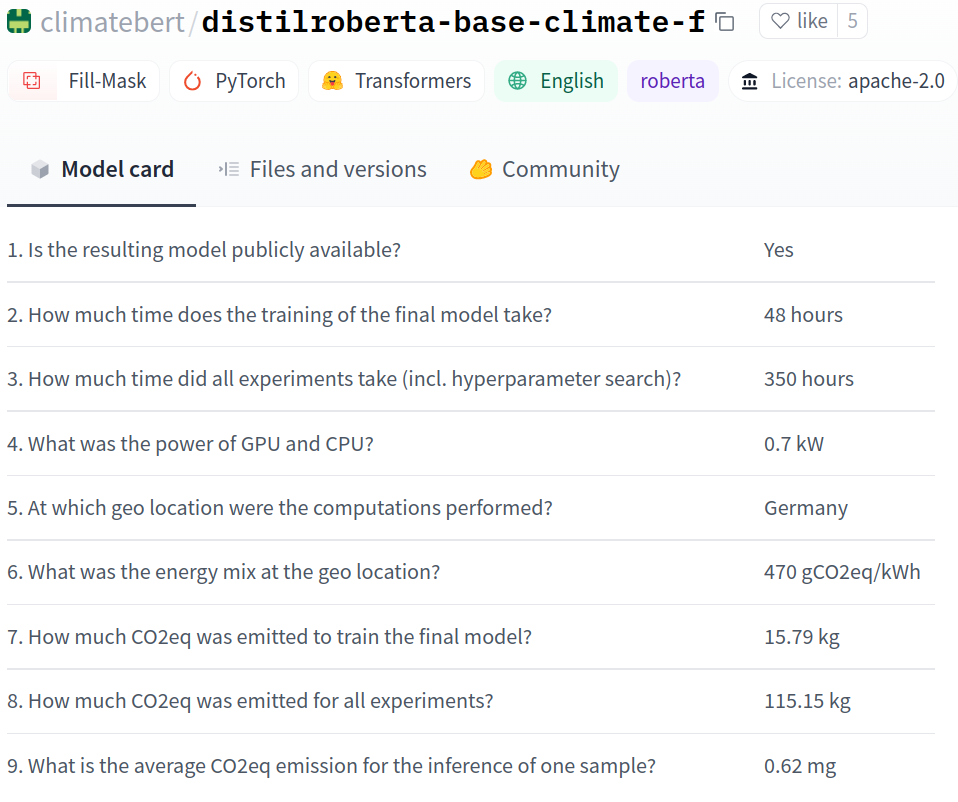}
    \caption{Example usage of our proposed climate performance model card (\S\ref{sec:model_cards}), on the Hugging Face Hub.\footnotemark}
    \label{fig:example_model_card}
\end{figure}
\footnotetext{\url{https://huggingface.co/climatebert/distilroberta-base-climate-f}}

This problem has been recognized by the NLP community, and a group of NLP researchers has recently proposed a policy document\footnote{\url{https://www.aclweb.org/portal/content/efficient-nlp-policy-document}} of recommendations for efficient NLP, aiming to minimize the greenhouse gas (GHG) emissions\footnote{GHG, CO$_2$, and carbon are used interchangeably in this paper. CO$_2$eq (or CO$_2$e), i.e., carbon dioxide equivalent translates GHG other than CO$_2$ into CO$_2$ equivalents based on the global warming potential \cite{brander2012greenhouse}.} resulting from experiments done as part of the research. 
This proposal is part of a research stream aiming towards \textit{Green NLP} and \textit{Green AI} \cite{Schwartz:2020}, which refers to ``AI research that yields novel results while taking into account the computational cost, encouraging a reduction in resources spent.''

While the branding of NLP and AI research as \textit{green} has raised some awareness of the environmental impact, the large majority of NLP researchers are still not aware of their environmental impact resulting from training and running of large computational models. This also explains why a research stream in a similar direction (see \S\ref{sec:automating}), in which software tools are proposed to measure carbon footprint while training models \citep{lacoste2019quantifying, henderson2020towards, anthony2020carbontracker, lottick2019energy}, have not been adopted by the community to a large extent (see \S\ref{sec:survey}). However, we claim that climate awareness is essential enough to be promoted in mainstream NLP (rather than only as a niche field) and that positive impact must be an inherent part of the discussion \cite{rolnick2019tackling,stede-patz-2021-climate}. Ideally environmental impact should always be taken into consideration, when deciding on which experiments to carry out.


We aim to simplify climate performance reporting in NLP while at the same time increasing awareness to its intricacies. Our contributions are:
\begin{itemize}
    \item We conduct a survey of environmental impact statements in NLP literature published in the past six years (\S\ref{sec:survey}). This survey is conducted across five dimensions that directly influence the environmental impact.
    \item We delineate the different notions of ``efficiency'' common in the literature, proposing a taxonomy to facilitate transparent reporting, and identify ten simple dimensions across which researchers can describe the environmental impact resulted by their research (\S\ref{sec:best_practices}).
    \item We propose a climate performance model card (\S\ref{sec:model_cards}) with the main purpose of being practically usable with only limited information about experiments and the underlying computer hardware (see Figure~\ref{fig:example_model_card}).
\end{itemize}

\section{Background}

\subsection{Automating Reporting}\label{sec:automating}
Several tools automate measurement and reporting of energy usage and emissions in ML.
\citet{lacoste2019quantifying} introduced a simple online calculator\footnote{\url{https://mlco2.github.io/impact/}} to estimate the amount of carbon emissions produced by training ML models. It can estimate the carbon footprint of GPU compute by manually specifying hardware type, hours used, cloud provider, and region.
\citet{henderson2020towards} presented a Python package\footnote{\url{https://github.com/Breakend/experiment-impact-tracker}} for consistent, easy, and more accurate reporting of energy, compute, and carbon impacts of ML systems by estimating them and generating standardized ``Carbon Impact Statements.''
\citet{anthony2020carbontracker} proposed a Python package\footnote{\url{https://github.com/lfwa/carbontracker}} that also has predictive capabilities, and allows proactive and intervention-driven reduction of carbon emissions. Model training can be stopped, at the user's discretion, if the predicted environmental cost is exceeded. 
\citet{victor_schmidt_2022_6369324} actively maintain a Python package\footnote{\url{https://codecarbon.io}} that, besides estimating impact and generating reports, shows developers how they can lessen emissions by optimizing their code or by using cloud infrastructure in geographical regions with renewable energy sources.
\citet{bannour-etal-2021-evaluating} surveyed and evaluated these tools and others for an NLP task, finding substantial variation in the reported measures due to different assumptions they make.
In summary, automated tools facilitate reporting, but they do not substitute awareness and should not be trusted blindly.

\subsection{Greenwashing}
While branding NLP and AI research as \textit{green} increases awareness of the environmental impact, there is a risk that the current framing, which exclusively addresses efficiency, will be perceived as the solution to the problem.
Of course, we attribute benevolent motives to the authors of the proposed policy document. Nevertheless, we would like to avoid a situation analogous to a common phenomenon in the financial field, where companies brand themselves as \textit{green} or \textit{sustainable} for branding or financial reasons, without implementing proportional measures in practice to mitigate the negative impact on the environment \cite{delmas2011drivers}. This malpractice is analogous to \textit{greenwashing}. While this is a general term, one aspect of greenwashing is ``a claim suggesting that a product is green based on a narrow set of attributes without attention to other important environmental issues'' \cite{choice2010sins}.\footnote{Paper, for example, is not necessarily environmentally preferable just because it comes from a sustainably harvested forest. Other important environmental issues in the paper-making process, such as greenhouse gas emissions or chlorine use in bleaching, may be equally important. Other examples are energy, utilities, and gasoline corporations that advertise about the benefits of new sources of energy while some are drilling into unexplored areas to source oil and thus destroying natural habitats and losing biodiversity, disguising the imbued hidden impacts \cite{de2020concepts}.} Our motivation is in line with the EU Commission's initiative to ``require companies to substantiate claims they make about the environmental footprint of their products/services by using standard methods for quantifying them.''\footnote{\url{https://ec.europa.eu/info/law/better-regulation/have-your-say/initiatives/12511-Environmental-performance-of-products-&-businesses-substantiating-claims_en}}
While \citet{Schwartz:2020} certainly do not argue that efficiency is \textit{sufficient} for sustainability, this notion, which is potentially implied by the \textit{green} branding, is misleading and even harmful: regardless of the extent of reduction, resources are still consumed, and GHGs are still emitted, among other negative effects. The efficiency mindset aims, at best, to prolong the duration of this situation. However, scaling up the performance of AI to satisfy the increasing demands from consumers risks ignoring the externalities incurred. Concepts such as reciprocity with the environment, which are central in some indigenous worldviews \cite{kimmerer2013braiding}, are absent from the discourse.

\subsection{Carbon Offsetting}
A common perception is that \textit{carbon neutrality} can be achieved by compensating for emissions by financial contributions, a practice referred to as \textit{carbon offsets}. This approach is problematic and controversial: the level of carbon prices required to achieve climate goals is highly debated \cite{https://doi.org/10.1002/wcc.207}. The Intergovernmental Panel on Climate Change (IPCC) and various international organizations like the International Energy Agency (IEA) clearly state that mitigation activities are essential. Compensation activities will be necessary for hard-to-abate-sectors, once all other technological solutions have been implemented, and where mitigation is not (yet) feasible.\footnote{See for example \url{www.ipcc.ch/sr15/} and \url{www.iea.org/reports/world-energy-outlook-2021}.} Moreover, economic dynamic efficiency requires investments in decarbonization technologies to keep the climate targets within reach. Compensation activities, especially in the afforestation area, delay the needed investments. This delay might exacerbate the likelihood of crossing climate tipping points and/or yields to a disorderly transition to a decarbonized economy \citep{esrb2016}.

\section{Survey of Climate Discussion in NLP}\label{sec:survey}

The issue of environmental impact is more general and not limited to NLP, but relevant to the entire field of AI: \citet{Schwartz:2020} surveyed papers from ACL, NeurIPS, and CVP. They noted whether authors claim their main contribution to improving accuracy or some related measure, an improvement to efficiency, both, or other. In all the conferences they considered, a large majority of the papers target accuracy.
However, we claim that the issue is more complex, and it is not sufficient to consider only the ``main contribution.'' Every paper should ideally have a positive impact or provide sufficient information to discuss meaningful options to reduce and mitigate negative impacts.

\begin{figure}[t]
	\centering
    \includegraphics[width=\columnwidth]{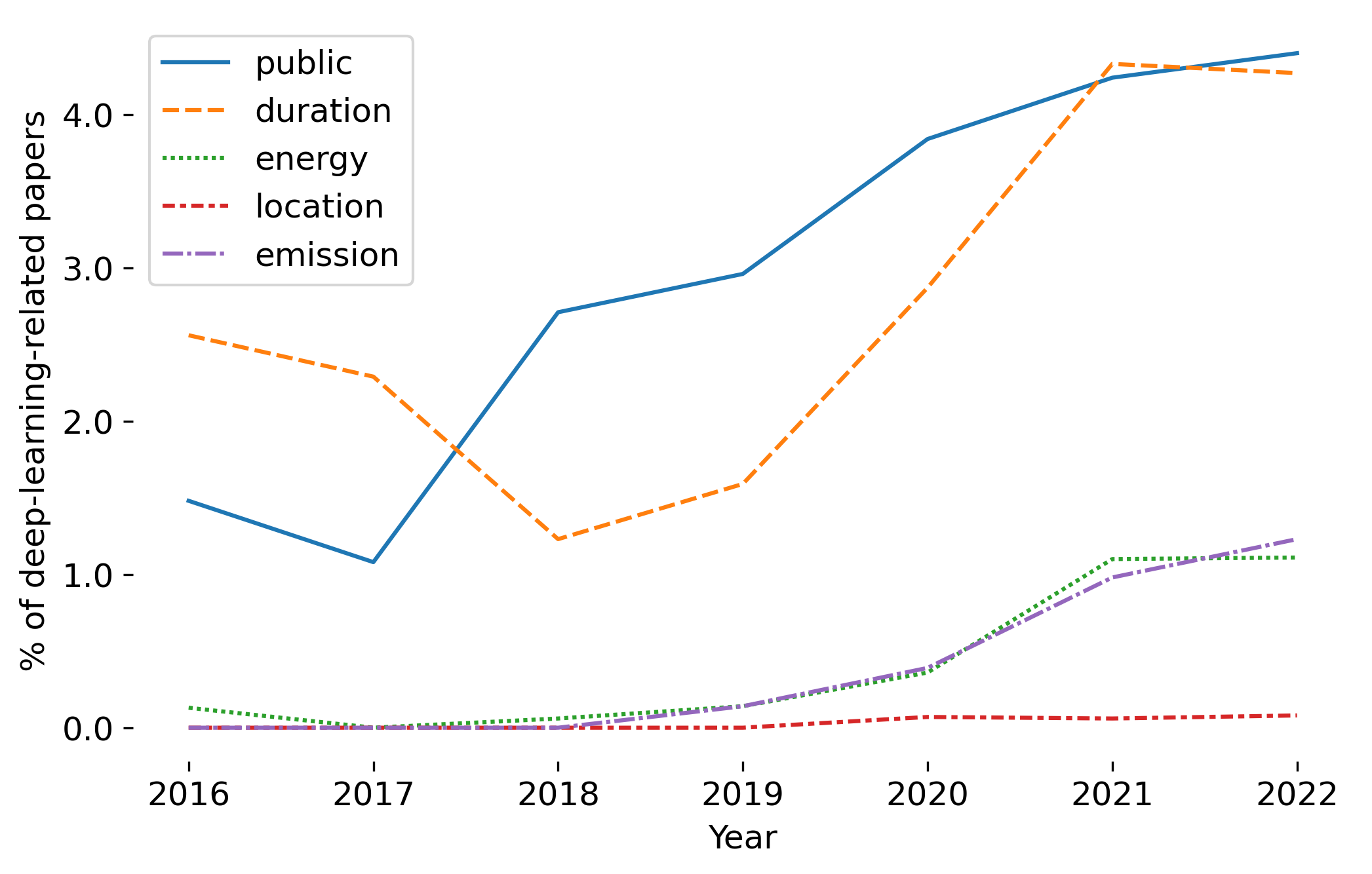}
	\caption{Development of proportions of deep-learning-related *ACL papers discussing \textit{public} model weights, \textit{duration} of model training or optimization, \textit{energy} consumption, \textit{location} where computations where performed, and \textit{emission} of GHG. While climate awareness is on the rise, it is still low overall. The numbers were calculated by counting pattern matches for papers in the ACL Anthology.}
	\label{fig:survey_proportions}
\end{figure}

\subsection{Quantitative Analysis}\label{sec:quantitative}
We analyze the statistics of papers in *ACL venues from 2016--2022 by downloading them from the ACL Anthology.\footnote{\url{https://aclanthology.org/}} However, instead of focusing on the main contribution, we look for any discussions on climate-related issues. We identify five dimensions in our study sample and create a regular expression pattern to match text for each (see Appendix~\ref{sec:patterns}).
These dimensions are public model weights, duration of model training or optimization, energy consumption, the location where computations are performed, and GHG emissions. If the pattern matches the text of a paper at least once,\footnote{While more recent papers have a dedicated ``broader impact'' section (see \S\ref{sec:model_cards}), in older papers, this discussion can appear anywhere in the text.} we consider that paper as discussing the corresponding category. We derive the proportions of papers by dividing the number of papers discussing a category by the number of deep learning-related papers. We only consider deep learning-related papers, as for these papers, climate-related issues are of much higher relevance than for those using other approaches \cite{strubell-etal-2019-energy}.

Figure~\ref{fig:survey_proportions} shows our findings. In general, researchers discuss climate-related issues more and more in their work. For instance, the proportion of papers that publish their model weights has almost quadrupled from about 1\% in 2017 to more than 4\% in 2022. We also find an increase in the proportion of papers that provide information on emissions or energy consumption. Nevertheless, the proportion for these categories remains low.

\begin{table}[t]
    \scriptsize
    \begin{tabular}{l|ccccc}
    \toprule
        Dimension & public & duration & energy & location & emission \\
        Proportion (\%) & 13 & 28 & 0 & 3 & 0 \\
    \bottomrule
    \end{tabular}
    \caption{Proportions along dimensions from Figure~\ref{fig:survey_proportions} in a manual annotated sample from EMNLP 2021.}
    \label{tab:sample_proportions}
\end{table}

\subsection{Manual Annotation}\label{sec:manual}
To complement our automatic pattern-based search approach, we also manually annotate a random sample of 100 papers from EMNLP 2021\footnote{\url{https://2021.emnlp.org/}} for the same five dimensions as before. Table~\ref{tab:sample_proportions} shows the proportions. Borderline cases are counted as ``reported'' for an optimistic estimate. This leads to the proportions of papers publishing model weights (13\%) and the duration of model training or optimization (28\%) being much higher than with our automatic approach, which cannot judge borderline cases and is thus more restrictive. Still, these proportions are at a low level. The proportion of papers reporting on the location, energy consumed and GHG emitted are in line with the results from our pattern-based search.

\subsection{Qualitative Survey}\label{sec:qualitative}
We examine article contents to ensure precision and to elaborate on existing practices. We review papers reporting on at least one dimension according to our pattern-based search or our manual analysis. Interestingly, many papers provide information in the context of reproducibility, publishing code but not necessarily model weights and reporting computation time only for specific steps. 

As examples for specific papers that go beyond what is usually expected in terms of reporting,
\citet{anderson-gomez-rodriguez-2021-modest} evaluate both accuracy and efficiency in dependency parsers, finding that different approaches are preferable depending on whether accuracy, training time or inference time are prioritized.
\citet{lakim-etal-2022-holistic} provide a detailed holistic assessment of the carbon footprint of an Arabic language model, considering the entire project, including data storage, researcher travel, training and deployment.

Our findings highlight the need to raise awareness of climate-related issues further and find a simple but effective way to report them transparently. Besides awareness and facilitation, incentives to address these issues could be a complementary approach. However, in the rest of this paper we focus on the former ``intrinsic'' motivation factors, leaving ``extrinsic'' motivation factors to future work.

\section{Towards Actionable Awareness}\label{sec:best_practices}

Efficiency (alongside accuracy) has been one of the main objectives in NLP (and computer science in general) long before its environmental aspects have been widely considered. In general, it refers to the amount of resources consumed (input) in order to achieve a given goal, such as a specific computation or accuracy in a task (output). Different definitions of efficiency correspond to different concepts of input and output.
It is crucial to (1) understand the different concepts, (2) be aware of their differences and consequently their climate impact, and (3) converge towards a set of efficiency measures that will be applied for comparable climate performance evaluation in NLP research.

\subsection{Related Work in NLP and AI}

\citet{strubell-etal-2019-energy} quantify the financial and environmental cost of various NLP models, exposing substantial costs from model development and not just final model training. They recommend reporting training time and hyperparameter sensitivity, and prioritizing efficient hardware and algorithms.

\citet{Schwartz:2020} compare several efficiency measures, focusing on input or resource consumption: CO$_2$eq emission, electricity usage, elapsed real time, number of parameters, and FPO (floating-point operations). They suggest FPO as a concrete, reliable measure for climate-related efficiency that does not depend on the underlying hardware, local electricity infrastructure, or algorithmic details. They suggest measuring efficiency as a trade-off between performance and training set size to enable comparisons with small training budgets.

\citet{henderson2020towards} show that FPOs ``are not adequate on their own to measure energy or even runtime efficiency.'' They recommend reporting various key figures, providing an automatic tool.

Alongside improvements in measurement methods, AI computations increasingly utilize cloud infrastructures, hindering transparency. \Citet{dodge2022measuring} provide a framework to measure carbon intensity in cloud instances, finding that data center region and time of day play significant roles.

Finally, \citet{MLSYS2022_ed3d2c21} highlight the role of system hardware development in AI environmental impact, encouraging a holistic perspective.




\subsection{Adopting Principles from Finance}\label{sec:principles}

The Greenhouse Gas Protocol\footnote{\url{https://ghgprotocol.org/}} is a widely used reporting framework for corporates. However, this standard does not foresee, so far, an explicit ICT (information and communications technology) component. We build on the general principles of the GHG Protocol (relevance, completeness, consistency, transparency, and accuracy) to propose principles for improving climate-related performance reporting of AI. While the Greenhouse Gas Protocol focuses on GHG emissions, we propose a more general framework corresponding to the different concepts of efficiency. We, therefore, replace the term GHG emissions with the term \textit{climate-related performance assessments}.\footnote{While the primary focus is about eventual GHG emissions in our case as well, by addressing \textit{climate-related performance}, we shift the focus from their direct measurement to a more holistic viewpoint.}

\begin{description}
    \item[Relevance] Ensure the climate-related performance assessment appropriately reflects the climate-related performance of training, evaluation and deployment, and serves the decision-making needs of users---both internal and external to the research group.
    Consider both factors inherent to the model (e.g., number of parameters) and model-external factors (e.g., energy mix).
    \item[Completeness] Account for and report on all relevant climate-related performance assessment items, using standardized model cards (see \S\ref{sec:model_cards}) to ensure accessibility to relevant information. Disclose and justify any specific exclusions or missing information, and explain which data input would be required to provide it. State how you will deal with the missing information in the future to reduce information gaps.
    \item[Consistency] Use consistent methodologies to make meaningful comparisons of reported emissions over time. Transparently document any changes to the data, inventory boundary, methods, or other relevant factors in the time series. Use readily-available emission calculation tools to ease comparison with other models. If you decide not to use available tools, explain why you deviate from available tools and report your assumptions about the energy mix, the conversion factors, and further assumptions required to calculate model-related emissions.
    \item[Transparency] Address all relevant issues factually and coherently to allow reproducible measurement of climate-related performance by independent researchers. Disclose any relevant assumptions and refer to the accounting and calculation methodologies and data sources used.
    \item[Accuracy of reporting] Achieve sufficient accuracy of the quantification of climate-related performance to enable users to make decisions with reasonable assurance as to the integrity of the reported information. Ensure that you report on the climate-related performance, even if you are in doubt about the accuracy. If in doubt, state the level of confidence.
\end{description}

\subsection{Actions Towards Improvement}

Reporting climate-related performance is not a goal on its own. Instead, it should be a means to raise awareness and translate it into actionable climate-related performance improvements when training and deploying a model. In addition, climate-aware model performance evaluations should ensure that downstream users of the technology can use the model in a climate-constrained future. Researchers should aim for climate-resilient NLP and algorithms to unlock long-term positive impacts. How to future-proof AI and NLP models should become an essential consideration in setting up any project.

The overall process of integrating these considerations would use enhanced transparency to unlock actionable awareness. Reporting on climate-related model performance should put researchers in a position to reflect on their setup and take immediate action when training the next model. To support this reflection for the researchers, the following proposes our climate performance model card.

\section{Climate Performance Model Cards}\label{sec:model_cards}

Since 2020, NeurIPS requires all research papers to submit broader impact statements
\cite{castelvecchi2021prestigious,gibney2020battle}. NLP conferences followed suit and introduced optional ethical and impact statements, starting with ACL in 2021.\footnote{See, e.g., the ACL Rolling Review Responsible NLP Research checklist: \url{https://aclrollingreview.org/responsibleNLPresearch/}.}
\citet{leins-etal-2020-give} discuss what an ethics assessment for ACL should look like but focus solely on political and societal issues.
\citet{tucker2020social} analyze the implications of improved data efficiency in AI but only discuss the societal aspect of access in research and industry, leaving environmental issues unexplored.
\citet{10.1145/3287560.3287596} introduced model cards to increase transparency about data use in AI, similarly due to societal issues. We propose extending impact statements and model cards to include information about the climate-related performance of the development and training of the model, improvements compared to alternative solutions, measures undertaken to mitigate negative impact, and importantly, about the expected climate-related performance of reusing the model for research and deployment.\footnote{See Appendix~\ref{sec:example_model_card} for a detailed example of the filled out model card from Figure~\ref{fig:example_model_card}.}

Our proposed model card also includes any \textit{positive} impact on the environment. A large direct negative impact does not rule out net positive impact due to contribution to downstream environmental efforts. While net impact cannot be measured objectively, since it depends on priorities and projections on the future use of the technology, we can set a framework for discussing this complex issue, providing researchers with the best practices to inform future researchers and practitioners.

\begin{table}[t]
\small
\adjustbox{width=\columnwidth}{
\begin{tabular}{@{}p{6cm}p{1cm}@{}}
\toprule
\multicolumn{2}{c}{\textbf{Minimum card}} \\
\midrule
\textbf{Information} & \textbf{Unit} \\
\midrule
1. Is the resulting model publicly available? & Yes/No \\
2. How much time does the training of the final model take? & Time \\
3. How much time did all experiments take (incl. hyperparameter search)? & Time \\
4. What was the power of GPU and CPU? & Watt \\
5. At which geo location were the computations performed? & Location \\\\
\midrule
\multicolumn{2}{c}{\textbf{Extended card}} \\
\midrule
6. What was the energy mix at the geo location? & gCO$_2$eq/ kWh \\
7. How much CO$_2$eq was emitted to train the final model? & kg \\
8. How much CO$_2$eq was emitted for all experiments? & kg \\
9. What is the average CO$_2$eq emission for the inference of one sample? & kg \\
10. Which positive environmental impact can be expected from this work? & Notes \\
11. Comments & Notes \\
\bottomrule
\end{tabular}
}
\caption{Proposed climate performance model card.}
\label{tbl:model_card}
\end{table} 

Table~\ref{tbl:model_card} shows our proposed sustainability model card, structured into a minimum card and an extended card. The minimum card contains very basic information about the distribution of the model, its purpose for the community, and roughly the computational work that has been put into the optimization of the models. The extended card then includes the energy mix to compute the CO$_2$eq emissions. In total, our sustainability model card contains eleven elements:
\begin{enumerate}
    \item \textbf{Publicly available artefacts.} In recent years, NLP researchers often make their final model available for the public. This trend came up to increase transparency and reprehensibility, yet, at the same time, it avoids the necessity to train frequently used models multiple times across the community \cite{wolf-etal-2020-transformers}. Thus, by publishing model (weights), computational resources and thereby CO$_2$eq emissions can be reduced.
    \item \textbf{Duration---training of final model.} This field denotes the time it took to train the final model (in minutes/hours/days/weeks). In case, there are multiple final models, this field asks for the training time of the model which has been trained the longest.
    \item \textbf{Duration of all computations.} The duration of all computations required to produce the results of the research project is strongly correlated with the CO$_2$eq emissions. Thus, we want to motivate NLP researchers to vary model types and hyperparameters reasonably. While determining the beginning of a project and deciding what counts as an experiment are in many cases difficult and subjective, we claim that an estimate of this quantity, along with a transparent confidence margin, is better than leaving it unreported.
    \item \textbf{Power of hardware.} Besides the duration of training, the power of the main hardware is a driving factor for CO$_2$eq emissions. Depending on the implementation, the majority of energy is consumed by CPUs or GPUs. We ask researchers to report the power in watts of the main hardware being used to optimize the model. For the sake of simplicity, we ask to specify the peak power of the hardware, for which the sum of the thermal design power (TDP) of the individual hardware components is a reasonable proxy. The manufacturers provide this information e.g. on their website.\footnote{See, for instance, \url{https://www.nvidia.com/en-us/data-center/a100/\#specifications} or \url{https://ark.intel.com/content/www/us/en/ark.html}. Alternatively, users can run \texttt{nvidia-smi} on the command line if using an NVIDIA GPU.} We want to underline again, that this model card's objective is not to have the most precise information but rather to have a rough estimate about the power.
    \item \textbf{Geographical location.} The energy mix (the CO$_2$eq emissions per watt consumed) depends on the geographical location. Thus, it is important to report where the model was trained. 
    \item \textbf{Energy mix at geographical location.} To compute the exact CO$_2$eq emissions, the energy mix at the geographical location is required. Organizations such as the International Energy Agency (IEA)\footnote{See \url{https://www.iea.org/countries}.} report these numbers. 
    \item \textbf{CO$_2$eq emissions of the final model.} This field describes an estimation for the emitted CO$_2$eq. Given the time for the computation (see item 3), the power, and the energy mix, the total CO$_2$eq emissions for the research can be calculated by \begin{align*}&\text{ComputationTime (hours) } \times \\ &\text{Power (kW) } \times \\ &\text{EnergyMix (gCO$_2$eq/kWh)} = \\  &\text{gCO$_2$eq}.\end{align*}
    Although awareness of the factors that affect CO$_2$eq emissions is important, we recommend using automated tools for the actual calculation (see~\S\ref{sec:automating}).
    \item \textbf{Total CO$_2$eq emissions.} Similar to the previous item, this field describes the total CO$_2$eq emitted during the training of all models. The calculation is equivalent to item 8.
    \item \textbf{CO$_2$eq emissions for inference.} Given that a model might be deployed in the future, the expected CO$_2$eq emissions in use of the model can be of value. To assure comparison between models, we ask the authors to report the average CO$_2$eq emission for the inference of one sample. For a dataset of $n$ samples, it can be calculated by \begin{align*}&1/n \times \text{InferenceTime (hours) }\times \\ &\text{Power (kW) }\times \\ &\text{EnergyMix(gCO$_2$eq/kWh)} = \\  &\text{gCO$_2$eq}.\end{align*}
    \item \textbf{Positive environmental impact.} NLP technologies begin to mature to the point where they could have an even broader impact and support to address major problems such as climate change. In this field, authors can state the expected positive impact resulting from their research. In case that the underlying work is not likely to have a direct positive impact, authors can also categorize their work into ``fundamental theories'', ``building block tools'', ``applicable tools'', or ``deployed applications'' \cite{jin-etal-2021-good}, and discuss why their work could set the basis for future work with a positive environmental impact.
    \item \textbf{Comments.} The objective of this climate performance model card is to collect the most relevant information about the computational resources, energy consumed, and CO$_2$eq emitted that were the result of the conducted research. Comments can include information about whether a number is likely over- or underestimated. In addition, this field can be used to provide the reader with indications of possible improvements in terms of energy consumption and CO$_2$eq emissions.
\end{enumerate}

\section{Discussion}

AI and NLP research are behind in incorporating sustainability discourse in the discussion. In the field of finance, an increasing amount of companies worldwide are soon required to state their environmental and broader sustainability-related impacts and/or commitments in their annual reports, mostly following the recommendations laid out by the Task Force on Climate-related Financial Disclosures \cite[TCFD;][]{board2017task}.\footnote{For instance, in the United Kingdom a new legislation will require firms to disclose climate-related financial information, with rules set to come into force from April 2022.}

\paragraph{Responsibility and accountability.}
Significant differences exist between annual reports and research papers:
companies are increasingly asked to take responsibility for their actions and are held accountable to their commitments by stakeholders, while researchers can shake off responsibility by transferring it to practitioners who use technology based on their research. Researchers are thus never held responsible for committing to reducing negative environmental impact unless they choose to submit their work to specific workshops or conference tracks on sustainable and efficient NLP. However, there are no best practices on what they can do to help those who are responsible for committing to sustainability---what information is necessary for accurate reporting and informed decision making?

\paragraph{Extrapolation to indirect impact.}
The quantification of indirect impact during reuse and deployment of artifacts developed in research is complex and can only be estimated. We, therefore, expect that this discussion in environmental impact statements will be more abstract and harder to assess. 
As a framework, we propose borrowing the notion of scopes from corporate GHG accounting \citep{patchell2018can}, where scopes 1, 2 and 3 correspond, respectively, to direct emissions (not applicable to NLP research); indirect emissions from operations, e.g., due to energy consumption (very common in NLP research); and indirect emissions upstream or downstream the value chain.
For our case, we suggest the following scopes:
\begin{compactenum}
    \item Emissions generated during experiments for the paper itself, usually electricity consumption-related.
    \item Impact on other researchers and practitioners in reducing emissions using the technology.
    \item The use of the technology for reducing emissions or other positive impact.
\end{compactenum}
Note that these correspond, respectively, to scopes 2, 3 and 3 in the GHG Protocol mentioned above.

\paragraph{Multi-objective optimization.}
Performance should not only be assessed in terms of output, but also inputs required to obtain a certain outcome. Based on this principle, performance evaluation should be based on both model performance and climate performance (cf. Table~\ref{tbl:model_eval}).
This can take the form of explicitly introducing climate performance into the objective function for optimization \cite{puvis-de-chavannes-etal-2021-hyperparameter} and in benchmarking \cite{NEURIPS2021_55b1927f}.

\begin{table}[t]
\adjustbox{width=\columnwidth}{
\begin{tabular}{lp{6cm}}
\toprule
Standard & \textbf{Model performance} (i.e., model output accuracy) \\
Emerging & \textbf{Climate-related performance} (i.e., CO$_2$eq emissions generated by training, deploying and using the model) \\
Future & \textbf{Climate-related efficiency performance} (i.e., marginal accuracy improvements relative to marginal input requirements) \\
\bottomrule
\end{tabular}
}
\caption{Extended model performance evaluation.}
\label{tbl:model_eval}
\end{table} 


\paragraph{Positive impact.}
NLP is relevant in several aspects to the UN sustainable development goals \cite{vinuesa2020role,conforti-etal-2020-natural,swarnakar2021nlp}.
\citet{jin-etal-2021-good} defined a framework for the social impacts of NLP, of which environmental impacts are a special case.
They define an impact stack consisting of four stages, from (1) fundamental theory to (2) building block tools and (3) applicable tools, and finally to (4) deployed applications.
Furthermore, they identify questions related to sustainable development goals for which NLP is relevant.
They categorize Green NLP as relevant only to the particular goal of ``mitigating problems brought by NLP,'' by minimizing direct impact as part of technology development. However, we claim that Green NLP must be viewed more broadly.
For example, \citet{rolnick2019tackling} discuss how machine learning can be used to tackle climate change, listing several fields with identified potential. For NLP, they mention the impact on the future of cities, on crisis management, individual action (understanding personal footprints, facilitating behavior change), informing policy for collective decision-making, education, and finance.
\citet{stede-patz-2021-climate} note that the topic of climate change has received little attention in the NLP community and propose applying NLP to analyze the climate change discourse, predict its evolution and respond accordingly.
Indeed, NLP is increasingly being used to analyze sustainability reports and environmental claims, facilitating enforcement of reporting requirements \cite{luccioni2020analyzing,bingler2021cheap,stammbach2022dataset}.

\section{Recommendations}\label{sec:impact}

As pointed out by \citet{Schwartz:2020}, a comparison between research and researchers from various locations and with various prerequisites can be difficult. Therefore, we want to point out rules of thumb that, in our opinion, should be followed by the authors of papers, as well as from reviewers who assess the quality thereof. 

\textbf{Do} increase transparency. With our climate performance model card, we aim to provide guidelines that give concrete ideas on how to report energy consumption and CO$_2$eq emissions. Our model card, on purpose, still allows for flexibility so that authors can change it to their respective setup. In case of high CPU usage, the authors can simplify their calculation of energy consumption by only looking at the CPU power; in the case of the GPU, it can simply be based on the GPU. Our main goal is transparency for users and increased awareness for modelers and researchers. Hence, transparency is to be weighted over accuracy. 

\textbf{Do} use the model cards to enable research institutions and practitioners to report on their climate performance and GHG emissions. An increasing number of first-moving research labs and institutes have started to account for their GHG emissions from direct energy use and flying, and intend to include their ICT emissions \citep[e.g.,][]{zurich2020eth, zurich2020uzh}. However, harmonized approaches are still lacking. Use the model cards to road-test how far they could support your institutions' GHG and climate impact reporting.

\textbf{Do not} use our model card for assessing research quality. The value of research is often only clear months or years after publication. Thus, the ratio between emitted CO$_2$eq and contribution to the NLP community cannot be measured accurately. Additionally, the emitted CO$_2$eq depends on the hardware used for the computations. Researchers working with less energy-efficient hardware would have a disadvantage if the emitted CO$_2$eq were being used for assessing the quality.
However, considering the energy efficiency of model performance might indirectly reduce a Global North--South bias, given that access to computational power is not evenly distributed across the World. Hence, targeting energy efficiency and reducing the computational power required to train and run models might mitigate some concerns on the inequality of research opportunities.

\textbf{Do not} report your voluntary financial climate protection contributions as emission offsetting. While emission offsetting used to be hailed as an efficient way to reduce global greenhouse gas emissions, this notion had to be revised with updated climate science consensus, at the latest with the IPCC's Special Report on Global Warming of 1.5C from 2018 \citep{masson2018global}.
Related to this aspect, do not communicate relative (efficiency-related) improvements as absolute climate-related performance improvements. 

\textbf{Do not} use this model card to assess net climate-related impacts. AI as an enabler for higher-order effects, for example, for climate-neutral economies and societies, is an important topic, which is, however, not in our scope. Instead, our approach aims to increase transparency about \textit{every model's first order effects}, be the model designed for societal change (or any other higher-order effect) or not.

\section{Conclusion}\label{sec:conclusion}

We argued that branding efficient methods in NLP as \textit{green} or \textit{sustainable} is insufficient and that due to the importance of the issue, climate awareness must be promoted in mainstream NLP rather than only in niche areas. We conducted a survey of climate discussion in NLP papers and found that climate-related issues are increasingly being discussed but are still uncommon. We proposed actionable measures to increase climate awareness based on experience from the finance domain and finally proposed a model card focusing on reporting climate performance transparently, which we encourage NLP researchers to use in any paper.

While our discussion, survey, and recommendations are aimed towards the NLP community, much is applicable to other AI fields. Indeed, specific recommendations have been made for machine learning \cite{henderson2020towards,Patterson2022} and medical image analysis \cite{selvan2022carbon}, for example. 
Concurrently, \citet{kaack2022aligning} propose a system-level roadmap addressing both GHG emissions and use of AI for climate change mitigation holistically.
Our focus on NLP enabled us to be more specific about relevant modeling components in our model card, as they are commonly used in NLP work. Furthermore, framing our arguments within the discourse initiated in the NLP community allowed us to address the specific points raised in this discussion so far, and highlight specific avenues for positive impact.



\input{limitations}


\bibliography{anthology,custom}
\bibliographystyle{acl_natbib}

\clearpage
\appendix

\section{Patterns for Quantitative Analysis}\label{sec:patterns}

The following are the regular expression patterns applied to identify papers according to the dimensions described in \S\ref{sec:quantitative}.

Public model weights:
\begin{lstlisting}[breaklines=true]
(((model|weight) (will be|is)?|(models|weights) (will be|are)?) (public|available|upload|made available|made public|provided (at|under|on)))|((publish|upload) [a-zA-Z0-9, ]{0,20}(model(s)?|weight(s)?))|(make [a-zA-Z0-9, ]{0,20}(model(s)?|weight(s)?) (available|public))|(provide [a-zA-Z0-9, ]{0,20}(model(s)?|weight(s)?) (at|under|on))
\end{lstlisting}

Duration of model training or optimization:
\begin{lstlisting}[breaklines=true]
(((pre(-)?)?train(ing|ed)?|optimize|optimization|(fine(-)?)?tun(e|ed|ing)) ([a-zA-Z0-9, ]{0,20})(for|took|take(s)?) ([a-zA-Z0-9, ]{0,20})(seconds|minute|hour|day|week|month)+)|hours of computation
\end{lstlisting}

Energy consumption:
\begin{lstlisting}[breaklines=true]
(energy|power|electricity) (consumption|usage)|(is|of|at) [1-9]{1}[0-9]{2,5} (watt(s)?|(k)?w) | pue
\end{lstlisting}

Location where computations are performed:
\begin{lstlisting}[breaklines=true]
((data ?center|(a|the) cloud|(virtual|gpu) machine|computer cluster|hpc) (is )?(at|in) )|(cloud|azure|google|aws)([a-zA-Z0-9, ]{0,20})region
\end{lstlisting}

GHG emission:
\begin{lstlisting}[breaklines=true]
(co2(e|eq)?|ghg|carbon) (footprint|emission(s)?|emitted|offset(ting)?)
\end{lstlisting}

The patterns were applied to the full paper text (including abstract, main contents and appendices, ignoring capitalization) for deep-learning-related papers identified by matching the following pattern:
\begin{lstlisting}[breaklines=true]
deep learning|neural network|lstm|recurrent neural network|rnn|transformer|mlp|convolutional neural network|cnn|gpt
\end{lstlisting}

\section{Example Model Card}\label{sec:example_model_card}
\begin{table}[t]
\small
\adjustbox{width=\columnwidth}{
\begin{tabular}{@{}p{55mm}p{25mm}@{}}
\toprule
\multicolumn{2}{c}{\textbf{ClimateBert}} \\
\midrule
1. Model publicly available? & Yes \\
2. Time to train final model &  8 hours \\
3. Time for all experiments &  288 hours \\
4. Power of GPU and CPU & 0.7 kW \\
5. Location for computations & Germany \\
6. Energy mix at location & 470 gCO$_2$eq/kWh \\
7. CO$_2$eq for final model & 2.63 kg \\
8. CO$_2$eq for all experiments & 94.75 kg \\
9. Average CO$_2$eq for inference per sample & 0.62 mg \\
\bottomrule
\end{tabular}
}
\caption{Climate performance model card for ClimateBert \citep{webersinke2021climatebert}.}
\label{tbl:model_card_example}
\end{table}

Table~\ref{tbl:model_card_example} provides an example climate performance model card according to the guidelines proposed in this paper. The model is ClimateBert, a language model which was finetuned on climate-related text \citep{webersinke2021climatebert}. The same information is provided on Hugging Face, illustrated in Figure~\ref{fig:example_model_card}.
Further information about each field is provided in the following:
\begin{enumerate}
    \item All weights of the final model are publicly available on \url{https://huggingface.co/climatebert}. The paper proposes a fine-tuned language model on climate-related text. Thus, the proposed models are specific to a field and not task agnostic.
    \item The duration for optimizing the final model was around 8 hours. Note, that the paper proposes four final models but this field should only mention the optimization time for one model.
    \item In total, we estimate the duration for all computations to be 12 days (=288 hours). This estimation is likely pessimistic, i.e., the duration for all computations was likely lower. However, we want to point out again that this model card values transparency over accuracy.
    \item The main hardware used for training were 2 x NVIDIA RTX A5000 with each GPU taking 230 watts. We add another 120 watts for the remaining hardware which would not be required by our model card.
    \item The models were all trained on servers in Germany.
    \item The energy mix is roughly 470 gCO$_2$eq/kWh.\footnote{According to \url{umweltbundesamt.de/publikationen/entwi cklung-der-spezifischen-kohlendioxid-7}.}
    \item Calculating \begin{align*}&8\,\text{hours} \times 0.7\, \text{kW} \times 470\,\text{gCO$_2$eq/kWh}\end{align*} leads to 2.63kg CO$_2$eq emissons.
    \item Calculating \begin{align*}&288\,\text{hours} \times 0.7\, \text{kW} \times 470\,\text{gCO$_2$eq/kWh}\end{align*} leads to 94.75kg CO$_2$eq emissons.
    \item A pass of 100,000 samples through the proposed model took 0.187 hours on the same server (using a batch size of 512). We then calculate \begin{align*}&\frac{0.187}{100,000}\,\text{hours} \times 0.7\, \text{kW} \times 470\,\text{gCO$_2$eq/kWh} \\&= 0.62\, \text{mgCO$_2$eq}\end{align*} as the emission for the inference of one sample.
\end{enumerate}
\paragraph{Positive impact.} The proposed language model on its own does not directly have a positive environmental impact. However, it can be used to train more accurate NLP models on climate-related downstream tasks. For instance, question-answering systems for climate-related topics or greenwashing detectors could benefit from this pretrained language model. This work can therefore be categorized as a ``building block tools'' following \citet{jin-etal-2021-good}, as it supports the training of NLP models in the field of climate change and, thereby, have a positive environmental impact in the future.
\paragraph{Possible improvements.} Block pruning is a method which drops a large number of attention heads in transformer models while only decreasing model performance slightly \citep{lagunas2021block}. Thus, the number of weights after block-pruning is decreased considerably which, in turn, decreases the CO$_2$eq emissions. Very likely, this method would show the same effect on the proposed ClimateBert model.

\begin{figure*}[t]
	\centering
        \includegraphics[width=0.7\textwidth]{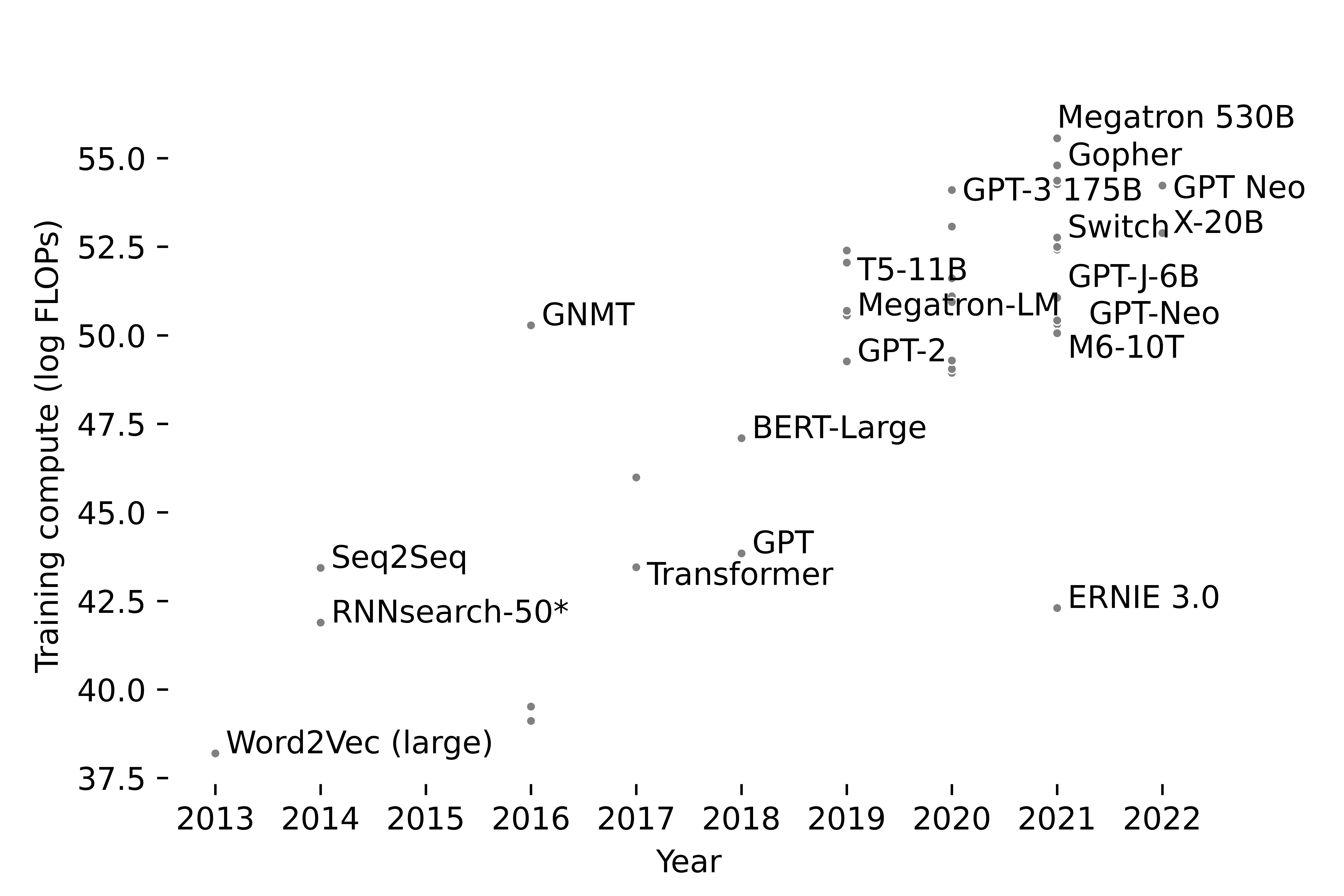}
	\caption{Development of floating point operations required to train NLP models. Note the log-scale.}
	\label{fig:flops_nlp}
\end{figure*}

\section{Timeline of Emissions in NLP}\label{sec:development}
Figure~\ref{fig:flops_nlp} shows the computational power that was put into the development of the major NLP models \citep{sevilla2022compute}. With few exceptions, the training compute for NLP models has steadily increased over the past decade. Although progress has also been made in terms of more energy efficient hardware (e.g., 19.5 GFLOPS/watt in a 2013 GTX Titan to 168.3 GFLOPs/watt in a 2021 RTX A6000), the increase in terms of required FLOPs is substantially larger. For example, going from GPT (in 2018) to GPT-3 175B (in 2020), the training compute increase from 1.1E19 to 3.14E23 FLOPs---an increase by a factor larger than 25,000.

\section{GHG Protocol Information Requirements for Companies}\label{sec:protocol}

Whilst the GHG Protocol does not provide an ICT sector tool, it provides emission factors by fuel source to calculate GHG emissions based on the energy consumption. The emission factors reflect the scientific climate consensus, based on the report of the Intergovernmental Panel on Climate Changes' latest Assessment Report---IPCC's AR5.\footnote{\url{https://www.ipcc.ch/report/ar5/syr/}. Note that AR6 will be released in late 2022 or early 2023: \url{https://www.ipcc.ch/ar6-syr/}.}

In terms of specific information to be disclosed, the GHG protocol guidance states several items relevant to NLP and ML research,\footnote{\url{https://ghgprotocol.org/corporate-standard}} which serve to build our model card approach. The items that can be used for our approach are presented in Figure~\ref{fig:ghg}.

\begin{figure*}[t]
\centering
\fbox{\parbox{.9\textwidth}{
\small
\begin{itemize}
    \item DESCRIPTION OF THE COMPANY AND INVENTORY BOUNDARY
    \begin{itemize}
    \item An outline of the organizational boundaries chosen, including the chosen consolidation approach
    \item An outline of the operational boundaries chosen, and if scope 3 is included, a list specifying which types of activities are covered.
    \end{itemize}
    \item INFORMATION ON EMISSIONS
    \begin{itemize}
        \item Total scope 1 and 2 emissions independent of any GHG trades such as sales, purchases, transfers, or banking of allowances.
        \item Emissions data separately for each scope. 
        \item Methodologies used to calculate or measure emissions, providing a reference or link to any calculation tools used. 
        \item Any specific exclusions of sources, facilities, and / or operations.
    \end{itemize}
    \item INFORMATION ON EMISSIONS AND PERFORMANCE
    \begin{itemize}
        \item Emissions data from relevant scope 3 emissions activities for which reliable data can be obtained. 
        \item Emissions data further subdivided, where this aids transparency, by business units/facilities, country, source types, and activity types 
        \item Relevant ratio performance indicators (e.g. emissions per kilowatt-hour generated, tonne of material production, or sales).
        \item An outline of any GHG management/reduction programs or strategies.
        \item An outline of any external assurance provided and a copy of any verification statement, if applicable, of the reported emissions data.
        \item Information on the quality of the inventory (e.g., information on the causes and magnitude of uncertainties in emission estimates) and an outline of policies in place to improve inventory quality. 
    \end{itemize}
    \item INFORMATION ON OFFSETS
    \begin{itemize}
        \item Information on offsets that have been purchased or developed outside the inventory boundary, subdivided by GHG storage/removals and emissions reduction projects. Specify if the offsets are verified/certified and/or approved by an external GHG program.
    \end{itemize}
\end{itemize}}}
\caption{Extract from the GHG Protocol Corporate Standard on which our climate performance reporting recommendations are based.}
\label{fig:ghg}
\end{figure*}

Furthermore, the GHG Protocols' Appendix A provides a guidance on accounting for indirect emissions from purchased electricity. This would be an important source of information for AI-related GHG accounting.

\end{document}

%% file: limitations.tex
\section{Limitations}\label{sec:limitations}
While climate awareness is necessary for including environmental considerations in decisions made during NLP research work, it is not sufficient for behavior change, namely, concrete actions by researchers and practitioners to reduce their negative impact and potentially contribute positively: as evident in various other societal issues, values do not necessarily determine behavior \cite{bostrom2020social}. Instead, climate-responsible behavior must also become ``the new normal'' for it to be mainstream. Awareness is only the first step in reaching that goal \cite{lockie2022mainstreaming}.

Furthermore, the climate awareness model card we propose requires less precise details than existing reporting tools \cite{lacoste2019quantifying,henderson2020towards,anthony2020carbontracker,victor_schmidt_2022_6369324}, which could limit its usefulness for informed decision making. However, as we claim in the paper, quantifying uncertainty may mitigate over-reliance on this information, which would otherwise possibly simply not have been reported at all.

Finally, if climate reporting becomes mandatory in NLP, it can actually be used for greenwashing if it entails financial or other incentives and if there is no control mechanism to check for honesty of the researchers. This is analogous to the situation in the corporate world, and can possibly be counteracted similarly, e.g., using ClimateBert.

%% file: green.bbl
\begin{thebibliography}{54}
\expandafter\ifx\csname natexlab\endcsname\relax\def\natexlab#1{#1}\fi

\bibitem[{Anderson and
  G{\'o}mez-Rodr{\'\i}guez(2021)}]{anderson-gomez-rodriguez-2021-modest}
Mark Anderson and Carlos G{\'o}mez-Rodr{\'\i}guez. 2021.
\newblock \href {https://doi.org/10.18653/v1/2021.iwpt-1.12} {A modest {P}areto
  optimisation analysis of dependency parsers in 2021}.
\newblock In \emph{Proceedings of the 17th International Conference on Parsing
  Technologies and the IWPT 2021 Shared Task on Parsing into Enhanced Universal
  Dependencies (IWPT 2021)}, pages 119--130, Online. Association for
  Computational Linguistics.

\bibitem[{Anthony et~al.(2020)Anthony, Kanding, and
  Selvan}]{anthony2020carbontracker}
Lasse F.~Wolff Anthony, Benjamin Kanding, and Raghavendra Selvan. 2020.
\newblock Carbontracker: Tracking and predicting the carbon footprint of
  training deep learning models.
\newblock ICML Workshop on Challenges in Deploying and monitoring Machine
  Learning Systems.
\newblock ArXiv:2007.03051.

\bibitem[{Bannour et~al.(2021)Bannour, Ghannay, N{\'e}v{\'e}ol, and
  Ligozat}]{bannour-etal-2021-evaluating}
Nesrine Bannour, Sahar Ghannay, Aur{\'e}lie N{\'e}v{\'e}ol, and Anne-Laure
  Ligozat. 2021.
\newblock \href {https://doi.org/10.18653/v1/2021.sustainlp-1.2} {Evaluating
  the carbon footprint of {NLP} methods: a survey and analysis of existing
  tools}.
\newblock In \emph{Proceedings of the Second Workshop on Simple and Efficient
  Natural Language Processing}, pages 11--21, Virtual. Association for
  Computational Linguistics.

\bibitem[{Bender et~al.(2021)Bender, Gebru, McMillan-Major, and
  Shmitchell}]{bender2021dangers}
Emily~M Bender, Timnit Gebru, Angelina McMillan-Major, and Shmargaret
  Shmitchell. 2021.
\newblock On the dangers of stochastic parrots: Can language models be too big?
\newblock In \emph{Proceedings of the 2021 ACM Conference on Fairness,
  Accountability, and Transparency}, pages 610--623.

\bibitem[{Bingler et~al.(2021)Bingler, Kraus, and Leippold}]{bingler2021cheap}
Julia~Anna Bingler, Mathias Kraus, and Markus Leippold. 2021.
\newblock Cheap talk and cherry-picking: What climate{B}ert has to say on
  corporate climate risk disclosures.
\newblock \emph{Available at SSRN}.

\bibitem[{Borgeaud et~al.(2022)Borgeaud, Mensch, Hoffmann, Cai, Rutherford,
  Millican, van~den Driessche, Lespiau, Damoc, Clark, de~Las~Casas, Guy,
  Menick, Ring, Hennigan, Huang, Maggiore, Jones, Cassirer, Brock, Paganini,
  Irving, Vinyals, Osindero, Simonyan, Rae, Elsen, and
  Sifre}]{borgeaud2022improving}
Sebastian Borgeaud, Arthur Mensch, Jordan Hoffmann, Trevor Cai, Eliza
  Rutherford, Katie Millican, George van~den Driessche, Jean-Baptiste Lespiau,
  Bogdan Damoc, Aidan Clark, Diego de~Las~Casas, Aurelia Guy, Jacob Menick,
  Roman Ring, Tom Hennigan, Saffron Huang, Loren Maggiore, Chris Jones, Albin
  Cassirer, Andy Brock, Michela Paganini, Geoffrey Irving, Oriol Vinyals, Simon
  Osindero, Karen Simonyan, Jack~W. Rae, Erich Elsen, and Laurent Sifre. 2022.
\newblock \href {http://arxiv.org/abs/2112.04426} {Improving language models by
  retrieving from trillions of tokens}.

\bibitem[{Bostr{\"o}m(2020)}]{bostrom2020social}
Magnus Bostr{\"o}m. 2020.
\newblock The social life of mass and excess consumption.
\newblock \emph{Environmental Sociology}, 6(3):268--278.

\bibitem[{Brander and Davis(2012)}]{brander2012greenhouse}
Matthew Brander and G~Davis. 2012.
\newblock Greenhouse gases, {CO}2, {CO}2e, and carbon: What do all these terms
  mean.
\newblock \emph{Econometrica, White Papers}.

\bibitem[{Castelvecchi et~al.(2021)}]{castelvecchi2021prestigious}
Davide Castelvecchi et~al. 2021.
\newblock Prestigious {AI} meeting takes steps to improve ethics of research.
\newblock \emph{Nature}, 589(7840):12--13.

\bibitem[{Conforti et~al.(2020)Conforti, Hirmer, Morgan, Basaldella, and
  Ben~Or}]{conforti-etal-2020-natural}
Costanza Conforti, Stephanie Hirmer, Dai Morgan, Marco Basaldella, and Yau
  Ben~Or. 2020.
\newblock \href {https://doi.org/10.18653/v1/2020.emnlp-main.677} {Natural
  language processing for achieving sustainable development: the case of neural
  labelling to enhance community profiling}.
\newblock In \emph{Proceedings of the 2020 Conference on Empirical Methods in
  Natural Language Processing (EMNLP)}, pages 8427--8444, Online. Association
  for Computational Linguistics.

\bibitem[{de~Freitas~Netto et~al.(2020)de~Freitas~Netto, Sobral, Ribeiro, and
  da~Luz~Soares}]{de2020concepts}
Sebasti{\~a}o~Vieira de~Freitas~Netto, Marcos Felipe~Falc{\~a}o Sobral, Ana
  Regina~Bezerra Ribeiro, and Gleibson~Robert da~Luz~Soares. 2020.
\newblock Concepts and forms of greenwashing: A systematic review.
\newblock \emph{Environmental Sciences Europe}, 32(1):1--12.

\bibitem[{Delmas and Burbano(2011)}]{delmas2011drivers}
Magali~A Delmas and Vanessa~Cuerel Burbano. 2011.
\newblock The drivers of greenwashing.
\newblock \emph{California management review}, 54(1):64--87.

\bibitem[{Devlin et~al.(2019)Devlin, Chang, Lee, and
  Toutanova}]{devlin-etal-2019-bert}
Jacob Devlin, Ming-Wei Chang, Kenton Lee, and Kristina Toutanova. 2019.
\newblock \href {https://doi.org/10.18653/v1/N19-1423} {{BERT}: Pre-training of
  deep bidirectional transformers for language understanding}.
\newblock In \emph{Proceedings of the 2019 Conference of the North {A}merican
  Chapter of the Association for Computational Linguistics: Human Language
  Technologies, Volume 1 (Long and Short Papers)}, pages 4171--4186,
  Minneapolis, Minnesota. Association for Computational Linguistics.

\bibitem[{Dodge et~al.(2022)Dodge, Prewitt, Tachet~des Combes, Odmark,
  Schwartz, Strubell, Luccioni, Smith, DeCario, and
  Buchanan}]{dodge2022measuring}
Jesse Dodge, Taylor Prewitt, Remi Tachet~des Combes, Erika Odmark, Roy
  Schwartz, Emma Strubell, Alexandra~Sasha Luccioni, Noah~A Smith, Nicole
  DeCario, and Will Buchanan. 2022.
\newblock Measuring the carbon intensity of {AI} in cloud instances.
\newblock In \emph{2022 ACM Conference on Fairness, Accountability, and
  Transparency}, pages 1877--1894.

\bibitem[{{ETH Zurich}(2021)}]{zurich2020eth}
{ETH Zurich}. 2021.
\newblock {ETH Zurich Sustainability Report 2019/2020}.
\newblock Technical report, ETH Zurich.

\bibitem[{{European Systemic Risk Board}(2016)}]{esrb2016}
{European Systemic Risk Board}. 2016.
\newblock \href {https://doi.org/doi/10.2849/703620} {\emph{Too late, too
  sudden : transition to a low-carbon economy and systemic risk}}.
\newblock European Systemic Risk Board.

\bibitem[{Fedus et~al.(2021)Fedus, Zoph, and Shazeer}]{fedus2021switch}
William Fedus, Barret Zoph, and Noam Shazeer. 2021.
\newblock Switch transformers: Scaling to trillion parameter models with simple
  and efficient sparsity.
\newblock \emph{arXiv preprint arXiv:2101.03961}.

\bibitem[{{Financial Stability Board}(2017)}]{board2017task}
{Financial Stability Board}. 2017.
\newblock Task force on climate-related financial disclosures.
\newblock \emph{Final Report: Recommendations of the Task Force on
  Climate-Related Financial Disclosures}.

\bibitem[{Gibney(2020)}]{gibney2020battle}
Elizabeth Gibney. 2020.
\newblock The battle for ethical {AI} at the world's biggest machine-learning
  conference.
\newblock \emph{Nature}, 577(7791):609--610.

\bibitem[{Henderson et~al.(2020)Henderson, Hu, Romoff, Brunskill, Jurafsky, and
  Pineau}]{henderson2020towards}
Peter Henderson, Jieru Hu, Joshua Romoff, Emma Brunskill, Dan Jurafsky, and
  Joelle Pineau. 2020.
\newblock Towards the systematic reporting of the energy and carbon footprints
  of machine learning.
\newblock \emph{Journal of Machine Learning Research}, 21(248):1--43.

\bibitem[{Hyams and Fawcett(2013)}]{https://doi.org/10.1002/wcc.207}
Keith Hyams and Tina Fawcett. 2013.
\newblock \href {https://doi.org/https://doi.org/10.1002/wcc.207} {The ethics
  of carbon offsetting}.
\newblock \emph{WIREs Climate Change}, 4(2):91--98.

\bibitem[{Jin et~al.(2021)Jin, Chauhan, Tse, Sachan, and
  Mihalcea}]{jin-etal-2021-good}
Zhijing Jin, Geeticka Chauhan, Brian Tse, Mrinmaya Sachan, and Rada Mihalcea.
  2021.
\newblock \href {https://doi.org/10.18653/v1/2021.findings-acl.273} {How good
  is {NLP}? a sober look at {NLP} tasks through the lens of social impact}.
\newblock In \emph{Findings of the Association for Computational Linguistics:
  ACL-IJCNLP 2021}, pages 3099--3113, Online. Association for Computational
  Linguistics.

\bibitem[{Kaack et~al.(2022)Kaack, Donti, Strubell, Kamiya, Creutzig, and
  Rolnick}]{kaack2022aligning}
Lynn~H Kaack, Priya~L Donti, Emma Strubell, George Kamiya, Felix Creutzig, and
  David Rolnick. 2022.
\newblock Aligning artificial intelligence with climate change mitigation.
\newblock \emph{Nature Climate Change}, pages 1--10.

\bibitem[{Kimmerer(2013)}]{kimmerer2013braiding}
Robin~Wall Kimmerer. 2013.
\newblock \emph{Braiding sweetgrass: Indigenous wisdom, scientific knowledge
  and the teachings of plants}.
\newblock Milkweed Editions.

\bibitem[{Lacoste et~al.(2019)Lacoste, Luccioni, Schmidt, and
  Dandres}]{lacoste2019quantifying}
Alexandre Lacoste, Alexandra Luccioni, Victor Schmidt, and Thomas Dandres.
  2019.
\newblock Quantifying the carbon emissions of machine learning.
\newblock \emph{Workshop on Tackling Climate Change with Machine Learning at
  NeurIPS 2019}.

\bibitem[{Lagunas et~al.(2021)Lagunas, Charlaix, Sanh, and
  Rush}]{lagunas2021block}
Fran{\c{c}}ois Lagunas, Ella Charlaix, Victor Sanh, and Alexander~M Rush. 2021.
\newblock Block pruning for faster transformers.
\newblock In \emph{Proceedings of the 2021 Conference on Empirical Methods in
  Natural Language Processing}, pages 10619--10629.

\bibitem[{Lakim et~al.(2022)Lakim, Almazrouei, Abualhaol, Debbah, and
  Launay}]{lakim-etal-2022-holistic}
Imad Lakim, Ebtesam Almazrouei, Ibrahim Abualhaol, Merouane Debbah, and Julien
  Launay. 2022.
\newblock \href {https://doi.org/10.18653/v1/2022.bigscience-1.8} {A holistic
  assessment of the carbon footprint of noor, a very large {A}rabic language
  model}.
\newblock In \emph{Proceedings of BigScience Episode {\#}5 -- Workshop on
  Challenges {\&} Perspectives in Creating Large Language Models}, pages
  84--94, virtual+Dublin. Association for Computational Linguistics.

\bibitem[{Leins et~al.(2020)Leins, Lau, and Baldwin}]{leins-etal-2020-give}
Kobi Leins, Jey~Han Lau, and Timothy Baldwin. 2020.
\newblock \href {https://doi.org/10.18653/v1/2020.acl-main.261} {Give me
  convenience and give her death: Who should decide what uses of {NLP} are
  appropriate, and on what basis?}
\newblock In \emph{Proceedings of the 58th Annual Meeting of the Association
  for Computational Linguistics}, pages 2908--2913, Online. Association for
  Computational Linguistics.

\bibitem[{Lockie(2022)}]{lockie2022mainstreaming}
Stewart Lockie. 2022.
\newblock Mainstreaming climate change sociology.

\bibitem[{Lottick et~al.(2019)Lottick, Susai, Friedler, and
  Wilson}]{lottick2019energy}
Kadan Lottick, Silvia Susai, Sorelle~A. Friedler, and Jonathan~P. Wilson. 2019.
\newblock Energy usage reports: Environmental awareness as part of algorithmic
  accountability.
\newblock \emph{Workshop on Tackling Climate Change with Machine Learning at
  NeurIPS 2019}.

\bibitem[{Luccioni et~al.(2020)Luccioni, Baylor, and
  Duchene}]{luccioni2020analyzing}
Sasha Luccioni, Emi Baylor, and Nicolas Duchene. 2020.
\newblock \href {https://www.climatechange.ai/papers/neurips2020/31} {Analyzing
  sustainability reports using natural language processing}.
\newblock In \emph{NeurIPS 2020 Workshop on Tackling Climate Change with
  Machine Learning}.

\bibitem[{Ma et~al.(2021)Ma, Ethayarajh, Thrush, Jain, Wu, Jia, Potts,
  Williams, and Kiela}]{NEURIPS2021_55b1927f}
Zhiyi Ma, Kawin Ethayarajh, Tristan Thrush, Somya Jain, Ledell Wu, Robin Jia,
  Christopher Potts, Adina Williams, and Douwe Kiela. 2021.
\newblock \href
  {https://proceedings.neurips.cc/paper/2021/file/55b1927fdafef39c48e5b73b5d61ea60-Paper.pdf}
  {Dynaboard: An evaluation-as-a-service platform for holistic next-generation
  benchmarking}.
\newblock In \emph{Advances in Neural Information Processing Systems},
  volume~34, pages 10351--10367. Curran Associates, Inc.

\bibitem[{Masson-Delmotte et~al.(2018)Masson-Delmotte, Zhai, P{\"o}rtner,
  Roberts, Skea, Shukla, Pirani, Moufouma-Okia, P{\'e}an, Pidcock
  et~al.}]{masson2018global}
Val{\'e}rie Masson-Delmotte, Panmao Zhai, Hans-Otto P{\"o}rtner, Debra Roberts,
  Jim Skea, Priyadarshi~R Shukla, Anna Pirani, W~Moufouma-Okia, C~P{\'e}an,
  R~Pidcock, et~al. 2018.
\newblock Global warming of 1.5 c.
\newblock \emph{An IPCC Special Report on the impacts of global warming of},
  1(5).

\bibitem[{Mitchell et~al.(2019)Mitchell, Wu, Zaldivar, Barnes, Vasserman,
  Hutchinson, Spitzer, Raji, and Gebru}]{10.1145/3287560.3287596}
Margaret Mitchell, Simone Wu, Andrew Zaldivar, Parker Barnes, Lucy Vasserman,
  Ben Hutchinson, Elena Spitzer, Inioluwa~Deborah Raji, and Timnit Gebru. 2019.
\newblock \href {https://doi.org/10.1145/3287560.3287596} {Model cards for
  model reporting}.
\newblock In \emph{Proceedings of the Conference on Fairness, Accountability,
  and Transparency}, FAT* '19, page 220–229, New York, NY, USA. Association
  for Computing Machinery.

\bibitem[{Patchell(2018)}]{patchell2018can}
Jerry Patchell. 2018.
\newblock Can the implications of the ghg protocol's scope 3 standard be
  realized?
\newblock \emph{Journal of Cleaner Production}, 185:941--958.

\bibitem[{Patterson et~al.(2022)Patterson, Gonzalez, Hölzle, Le, Liang,
  Munguia, Rothchild, So, Texier, and Dean}]{Patterson2022}
David Patterson, Joseph Gonzalez, Urs Hölzle, Quoc~Hung Le, Chen Liang,
  Lluis-Miquel Munguia, Daniel Rothchild, David So, Maud Texier, and Jeffrey
  Dean. 2022.
\newblock \href {https://doi.org/10.36227/techrxiv.19139645.v3} {{The Carbon
  Footprint of Machine Learning Training Will Plateau, Then Shrink}}.

\bibitem[{Puvis~de Chavannes et~al.(2021)Puvis~de Chavannes, Kongsbak, Rantzau,
  and Derczynski}]{puvis-de-chavannes-etal-2021-hyperparameter}
Lucas~H{\o}yberg Puvis~de Chavannes, Mads Guldborg~Kjeldgaard Kongsbak, Timmie
  Rantzau, and Leon Derczynski. 2021.
\newblock \href {https://doi.org/10.18653/v1/2021.sustainlp-1.12}
  {Hyperparameter power impact in transformer language model training}.
\newblock In \emph{Proceedings of the Second Workshop on Simple and Efficient
  Natural Language Processing}, pages 96--118, Virtual. Association for
  Computational Linguistics.

\bibitem[{Rolnick et~al.(2019)Rolnick, Donti, Kaack, Kochanski, Lacoste,
  Sankaran, Ross, Milojevic-Dupont, Jaques, Waldman-Brown, Luccioni, Maharaj,
  Sherwin, Mukkavilli, Kording, Gomes, Ng, Hassabis, Platt, Creutzig, Chayes,
  and Bengio}]{rolnick2019tackling}
David Rolnick, Priya~L. Donti, Lynn~H. Kaack, Kelly Kochanski, Alexandre
  Lacoste, Kris Sankaran, Andrew~Slavin Ross, Nikola Milojevic-Dupont, Natasha
  Jaques, Anna Waldman-Brown, Alexandra Luccioni, Tegan Maharaj, Evan~D.
  Sherwin, S.~Karthik Mukkavilli, Konrad~P. Kording, Carla Gomes, Andrew~Y. Ng,
  Demis Hassabis, John~C. Platt, Felix Creutzig, Jennifer Chayes, and Yoshua
  Bengio. 2019.
\newblock \href {http://arxiv.org/abs/1906.05433} {Tackling climate change with
  machine learning}.

\bibitem[{Schmidt et~al.(2022)Schmidt, Goyal-Kamal, Courty, Feld, SabAmine,
  Zhao, Joshi, Luccioni, Léval, Bogroff, Laskaris, LiamConnell, Wang, Catovic,
  Blank, Stęchły, alencon, Saboni, JPW, MinervaBooks, de~Lavoreille,
  McCarthy, Tae, Tourbier, and kraktus}]{victor_schmidt_2022_6369324}
Victor Schmidt, Goyal-Kamal, Benoit Courty, Boris Feld, SabAmine, Franklin
  Zhao, Aditya Joshi, Sasha Luccioni, Mathilde Léval, Alexis Bogroff, Niko
  Laskaris, LiamConnell, Ziyao Wang, Armin Catovic, Douglas Blank, Michał
  Stęchły, alencon, Amine Saboni, JPW, MinervaBooks, Hugues de~Lavoreille,
  Connor McCarthy, Jake Tae, Sébastien Tourbier, and kraktus. 2022.
\newblock \href {https://doi.org/10.5281/zenodo.6369324} {mlco2/codecarbon:
  v2.0.0a3}.

\bibitem[{Schwartz et~al.(2020)Schwartz, Dodge, Smith, and
  Etzioni}]{Schwartz:2020}
Roy Schwartz, Jesse Dodge, Noah~A. Smith, and Oren Etzioni. 2020.
\newblock \href {https://doi.org/10.1145/3381831} {Green {AI}}.
\newblock \emph{Communications of the ACM (CACM)}, 63(12):54--63.

\bibitem[{Selvan et~al.(2022)Selvan, Bhagwat, Anthony, Kanding, and
  Dam}]{selvan2022carbon}
Raghavendra Selvan, Nikhil Bhagwat, Lasse F~Wolff Anthony, Benjamin Kanding,
  and Erik~B Dam. 2022.
\newblock Carbon footprint of selecting and training deep learning models for
  medical image analysis.
\newblock \emph{arXiv preprint arXiv:2203.02202}.

\bibitem[{Sevilla et~al.(2022)Sevilla, Heim, Ho, Besiroglu, Hobbhahn, and
  Villalobos}]{sevilla2022compute}
Jaime Sevilla, Lennart Heim, Anson Ho, Tamay Besiroglu, Marius Hobbhahn, and
  Pablo Villalobos. 2022.
\newblock \href {http://arxiv.org/abs/2202.05924} {Compute trends across three
  eras of machine learning}.

\bibitem[{So et~al.(2022)So, Mańke, Liu, Dai, Shazeer, and Le}]{so2022primer}
David~R. So, Wojciech Mańke, Hanxiao Liu, Zihang Dai, Noam Shazeer, and
  Quoc~V. Le. 2022.
\newblock \href {http://arxiv.org/abs/2109.08668} {Primer: Searching for
  efficient transformers for language modeling}.

\bibitem[{Stammbach et~al.(2022)Stammbach, Webersinke, Bingler, Kraus, and
  Leippold}]{stammbach2022dataset}
Dominik Stammbach, Nicolas Webersinke, Julia~Anna Bingler, Mathias Kraus, and
  Markus Leippold. 2022.
\newblock \href {https://ssrn.com/abstract=4207369} {A dataset for detecting
  real-world environmental claims}.
\newblock \emph{Available at SSRN}.

\bibitem[{Stede and Patz(2021)}]{stede-patz-2021-climate}
Manfred Stede and Ronny Patz. 2021.
\newblock \href {https://doi.org/10.18653/v1/2021.nlp4posimpact-1.2} {The
  climate change debate and natural language processing}.
\newblock In \emph{Proceedings of the 1st Workshop on NLP for Positive Impact},
  pages 8--18, Online. Association for Computational Linguistics.

\bibitem[{Strubell et~al.(2019)Strubell, Ganesh, and
  McCallum}]{strubell-etal-2019-energy}
Emma Strubell, Ananya Ganesh, and Andrew McCallum. 2019.
\newblock \href {https://doi.org/10.18653/v1/P19-1355} {Energy and policy
  considerations for deep learning in {NLP}}.
\newblock In \emph{Proceedings of the 57th Annual Meeting of the Association
  for Computational Linguistics}, pages 3645--3650, Florence, Italy.
  Association for Computational Linguistics.

\bibitem[{Swarnakar and Modi(2021)}]{swarnakar2021nlp}
Pradip Swarnakar and Ashutosh Modi. 2021.
\newblock {NLP} for climate policy: Creating a knowledge platform for holistic
  and effective climate action.
\newblock \emph{arXiv preprint arXiv:2105.05621}.

\bibitem[{TerraChoice(2010)}]{choice2010sins}
TerraChoice. 2010.
\newblock The sins of greenwashing: home and family edition.
\newblock \emph{Underwriters Laboratories}.

\bibitem[{Tucker et~al.(2020)Tucker, Anderljung, and Dafoe}]{tucker2020social}
Aaron~D Tucker, Markus Anderljung, and Allan Dafoe. 2020.
\newblock Social and governance implications of improved data efficiency.
\newblock In \emph{Proceedings of the AAAI/ACM Conference on AI, Ethics, and
  Society}, pages 378--384.

\bibitem[{{UZH Zurich}(2021)}]{zurich2020uzh}
{UZH Zurich}. 2021.
\newblock {UZH Zurich Sustainability Report 2019/2020}.
\newblock Technical report, UZH Zurich.

\bibitem[{Vinuesa et~al.(2020)Vinuesa, Azizpour, Leite, Balaam, Dignum,
  Domisch, Fell{\"a}nder, Langhans, Tegmark, and Nerini}]{vinuesa2020role}
Ricardo Vinuesa, Hossein Azizpour, Iolanda Leite, Madeline Balaam, Virginia
  Dignum, Sami Domisch, Anna Fell{\"a}nder, Simone~Daniela Langhans, Max
  Tegmark, and Francesco~Fuso Nerini. 2020.
\newblock The role of artificial intelligence in achieving the sustainable
  development goals.
\newblock \emph{Nature communications}, 11(1):1--10.

\bibitem[{Webersinke et~al.(2021)Webersinke, Kraus, Bingler, and
  Leippold}]{webersinke2021climatebert}
Nicolas Webersinke, Mathias Kraus, Julia~Anna Bingler, and Markus Leippold.
  2021.
\newblock Climate{B}ert: A pretrained language model for climate-related text.
\newblock \emph{arXiv preprint arXiv:2110.12010}.

\bibitem[{Wolf et~al.(2020)Wolf, Debut, Sanh, Chaumond, Delangue, Moi, Cistac,
  Rault, Louf, Funtowicz, Davison, Shleifer, von Platen, Ma, Jernite, Plu, Xu,
  Le~Scao, Gugger, Drame, Lhoest, and Rush}]{wolf-etal-2020-transformers}
Thomas Wolf, Lysandre Debut, Victor Sanh, Julien Chaumond, Clement Delangue,
  Anthony Moi, Pierric Cistac, Tim Rault, Remi Louf, Morgan Funtowicz, Joe
  Davison, Sam Shleifer, Patrick von Platen, Clara Ma, Yacine Jernite, Julien
  Plu, Canwen Xu, Teven Le~Scao, Sylvain Gugger, Mariama Drame, Quentin Lhoest,
  and Alexander Rush. 2020.
\newblock \href {https://doi.org/10.18653/v1/2020.emnlp-demos.6} {Transformers:
  State-of-the-art natural language processing}.
\newblock In \emph{Proceedings of the 2020 Conference on Empirical Methods in
  Natural Language Processing: System Demonstrations}, pages 38--45, Online.
  Association for Computational Linguistics.

\bibitem[{Wu et~al.(2022)Wu, Raghavendra, Gupta, Acun, Ardalani, Maeng, Chang,
  Aga, Huang, Bai, Gschwind, Gupta, Ott, Melnikov, Candido, Brooks, Chauhan,
  Lee, Lee, Akyildiz, Balandat, Spisak, Jain, Rabbat, and
  Hazelwood}]{MLSYS2022_ed3d2c21}
Carole-Jean Wu, Ramya Raghavendra, Udit Gupta, Bilge Acun, Newsha Ardalani,
  Kiwan Maeng, Gloria Chang, Fiona Aga, Jinshi Huang, Charles Bai, Michael
  Gschwind, Anurag Gupta, Myle Ott, Anastasia Melnikov, Salvatore Candido,
  David Brooks, Geeta Chauhan, Benjamin Lee, Hsien-Hsin Lee, Bugra Akyildiz,
  Maximilian Balandat, Joe Spisak, Ravi Jain, Mike Rabbat, and Kim Hazelwood.
  2022.
\newblock \href
  {https://proceedings.mlsys.org/paper/2022/file/ed3d2c21991e3bef5e069713af9fa6ca-Paper.pdf}
  {Sustainable ai: Environmental implications, challenges and opportunities}.
\newblock In \emph{Proceedings of Machine Learning and Systems}, volume~4,
  pages 795--813.

\end{thebibliography}
